\documentclass[sigconf]{acmart}

\usepackage[noline,ruled,noend]{algorithm2e} 

\usepackage{enumitem}
\def\BibTeX{{\rm B\kern-.05em{\sc i\kern-.025em b}\kern-.08emT\kern-.1667em\lower.7ex\hbox{E}\kern-.125emX}}
    
\copyrightyear{2019}
\acmYear{2019}
\setcopyright{acmcopyright}
\acmPrice{15.00}

\begin{document}

%
\title{A Dynamic Evolutionary Framework for Timeline Generation based on Distributed Representations}

%
\author{Dongyun Liang, Guohua Wang, Jing Nie}
\affiliation{%
  \institution{Tencent}
}
\email{dylanliang@tencent.com}

%
\renewcommand{\shortauthors}{Dongyun Liang et al.}
\renewcommand{\shorttitle}{A Dynamic Evolutionary Framework for Timeline Generation}

%
\begin{abstract}
Given the collection of timestamped web documents related to the evolving topic,
timeline summarization (TS) highlights its most important events in the form of relevant summaries to represent the development of a topic over time.
Most of the previous work focuses on fully-observable ranking models and depends on hand-designed features or complex mechanisms that may not generalize well.
We present a novel dynamic framework for evolutionary timeline generation leveraging distributed representations, 
which dynamically finds the most likely sequence of evolutionary summaries in the timeline, called the Viterbi timeline, and reduces the impact of events that irrelevant or repeated to the topic.
The assumptions of the coherence and the global view run through our model. We explore adjacent relevance to constrain timeline coherence and make sure the events evolve on the same topic with a global view. 
Experimental results demonstrate that our framework is feasible to extract summaries for timeline generation, 
outperforms various competitive baselines, and achieves the state-of-the-art performance as an unsupervised approach.

\end{abstract}

%
%


%

%
\maketitle

\section{Introduction}

Along with the rapid growth of the World Wide Web, 
there is a huge document collection related to the news topics.
General search engines rank the indexed document according to their understanding of the user's query relevance.
and users have access to a complete range of ranked documents about a particular topic. 
However, users still possibly get lost in redundant results, even if the ranked documents have been ordered by time. 
It is significant to provide a timeline for users to view what evolutionary topic is going on and what key events break out along with a particular topic.

As handcrafted timeline requires tremendous human labor of reading and understanding, timeline summarization (TS) is a widely adopted task to generate timelines~\cite{yan2011timeline,li2013evolutionary,li2014timeline}. 
News topic can be broken into a sequence of events, and TS provides temporal summaries of the evolution of news events related to the topic.
Given news marked date, such as the indexed date by search engine in practice,
we aim to tackle this problem for selecting a subset of important dates as the major points along the timeline~\cite{kessler2012finding}, meanwhile, generating a good daily summary for these dates~\cite{tran2015timeline}.
Some researches~\cite{althoff2015timemachine,liu2017growing} use clustering and ranking techniques to select the important events that be included in the final summary, and most of the previous work relies on hand-designed features~\cite{chieu2004query,yan2011evolutionary} or complex mechanisms. 

In recent years, several distributed representations approaches \cite{mikolov2013efficient,bojanowski2017enriching} caused the trend of deep learning of the text.
More and more embeddings derived from deep network for word or sentence have been proved efficient in representing rich syntactic and semantic information, which can somewhat help people free from the tedious feature engineering.
There have been some efforts that explores this in TS~\cite{wang2016low,zhou2018neural}. 
Distributed representation of a target word or sentence derives from the association of adjacency~\cite{levy2014neural}, which carries around information to achieve semantic coherence and is conceptually similar to the coherence of adjacent event and the relevance on a same topic in TS.
However, as far as we know, most of the explorations in TS treat representations as a extra feature to enhance the effect~\cite{wang2016low}, there is no framework which is a natural use of distributed representations to integrate their adjacency relevance into the coherence of TS.

In this paper, we propose a novel framework to generate timeline, which is dynamic and evolutionary to make natural use of distributed representations.
In specific, we assume that the timeline has a certain coherence~\cite{yan2011evolutionary}.
That is, with the evolution of the topic, there is not necessarily a direct relevance between the first event and the last rather than the next event, but the adjacent events in the timeline have a certain relevance in the news reports. 
Since reporting a new event on a same topic often cites the recent event reports, coherent events always have partial similarities in text, and this effect will gradually weaken as the timeline develops.
We also assume that the timeline has a global view to guide the central theme~\cite{yan2011timeline}.
As the events are assigned to a particular topic, the events along the timeline are more or less related to this topic.

Our approach takes a dated collection relevant to a news topic as input, and output a timeline with the component summaries of events which represent evolutionary trajectories on specific dates.
Firstly, we perform embedding learning to model the continuous vector representations of the inputs.
Then we cluster massive amounts of news paragraphs into event groups according to their representations and the corresponding date, and we get several event groups in each step by date. 
Under the guidance of coherence and global view, the framework try to find a dynamic optimal path linking the events between these steps.
The experimental results show significant and consistent performance improvement over the state-of-the-art methods on public datasets.

\section{Related Work}

There have been many studies about timeline generation from
various sources, including the sentences, paragraphs or headlines of news.
\cite{chieu2004query} extracted the popular and bursting sentences to place along the timeline from a query. 
\cite{yan2011evolutionary} optimized the problem via iterating substitution by incorporating several constraints. 
Supervised learning~\cite{tran2013leveraging} is also widely used in TS, 
and \cite{yan2011timeline,wang2016socially} proposed a ranking framework to get temporal summarization.
Most of summarization corpora are text-only, and \cite{wang2016low} utilized both text and image to provide a comprehensive sketch of the topic evolution. 
Considering timeline as latent variables, many dynamic approaches~\cite{li2013evolutionary,liang2016dynamic} based on probabilistic graphical models have been proposed to discover the evolving patterns.
\cite{tran2013leveraging,tran2015timeline} published the datasets that consists of the timelines created by experts, the correlated news articles and headlines.

Some work~\cite{althoff2015timemachine,mishra2016event} has focused on a better understanding of a particular entity or event by displaying a list of episodes in time order, and they jointly consider the relevance and temporal diversity to
interpret the cause and effect of the entity.
Another related work is concerned on text stream summarization: \cite{liu2017growing} discovered key information in vast text, such as the events from trending and breaking news, then organized that. There exists explorations~\cite{li2014timeline} for individual timeline from Twitter. 
However, hand-designed features account for a large proportion in the above work, which may not generalize well. 
As deep learning has gained immense success on Natural Language Processing, ~\cite{wang2016low,zhou2018neural}  introduce distributed representations into TS.

\section{Framework}

\subsection{Preliminaries}
In our work, we focus on the news topic, such as \textit{2010 British Oil spill},  and the events that evolve with the development of topics. 
TS consists of a temporally ordered list of summaries, which describes the main events that occurred along the time.

Let $ D = \{D_1, D_2, \ldots, D_T\} $ is the set of news documents related to a particular topic $ q $, where $ D_i \in D $ is the subset of news collected on the period of $ i $-th day. 
We extract the paragraphs $ A_i = \{A_{i}^{1}, \ldots, A_{i}^{m}\} $ from $ D_i $ to denote the candidates of the TS , which can be the news headline, first sentence and n-gram sentences of the content.
$ TS_q = \{A_{t_1}^{m_1}, A_{t_2}^{m_2}, \ldots, A_{t_{|S|}}^{m_{|S|}} \} $ denotes the TS about $ q $, where 
$ A_{t_i}^{m_i} \in A_{t_i} $ is  the summary of the $ t_i $-th day in TS,
$ t_{i} \in \{1,\ldots , T \}$ is the specific date on the period of TS, $ m_i $ denotes the specific sample of $ A_{t_i} $ in $ t_i $-th day , and $ |S| $ is the total steps of the timeline.

\subsection{Learning Distributed Representations}

There are various methods to get distributed representations for short text. To learn the representations for the paragraphs, 
we filter stop words out of them, use skip-gram model~\cite{bojanowski2017enriching} to learn vector embeddings of the words, and multiply them with their TF-IDF scores to represent the paragraphs, which has been proven to be a baseline~\cite{kenter2016siamese} and feature across a multitude of tasks, especially short text similarity tasks.

Given the topic $ q $ and the dated news documents $ D $ related to $ q $ that can be obtained by the retrieval recall, 
we can optionally expand $q$ by some keywords or descriptive text, such as Wikipedia knowledge returned by search engine, and extract paragraphs subset $ A_i $ from $ D_i $. 
As mentioned above, $ q $ is embedded to a vector $ v(q) $, and $ A_{i}^{j}$ is embedded to a vector $ v(A_{i}^{j}) $.

\begin{figure}
\includegraphics[height=1.8in, width=3.2in]{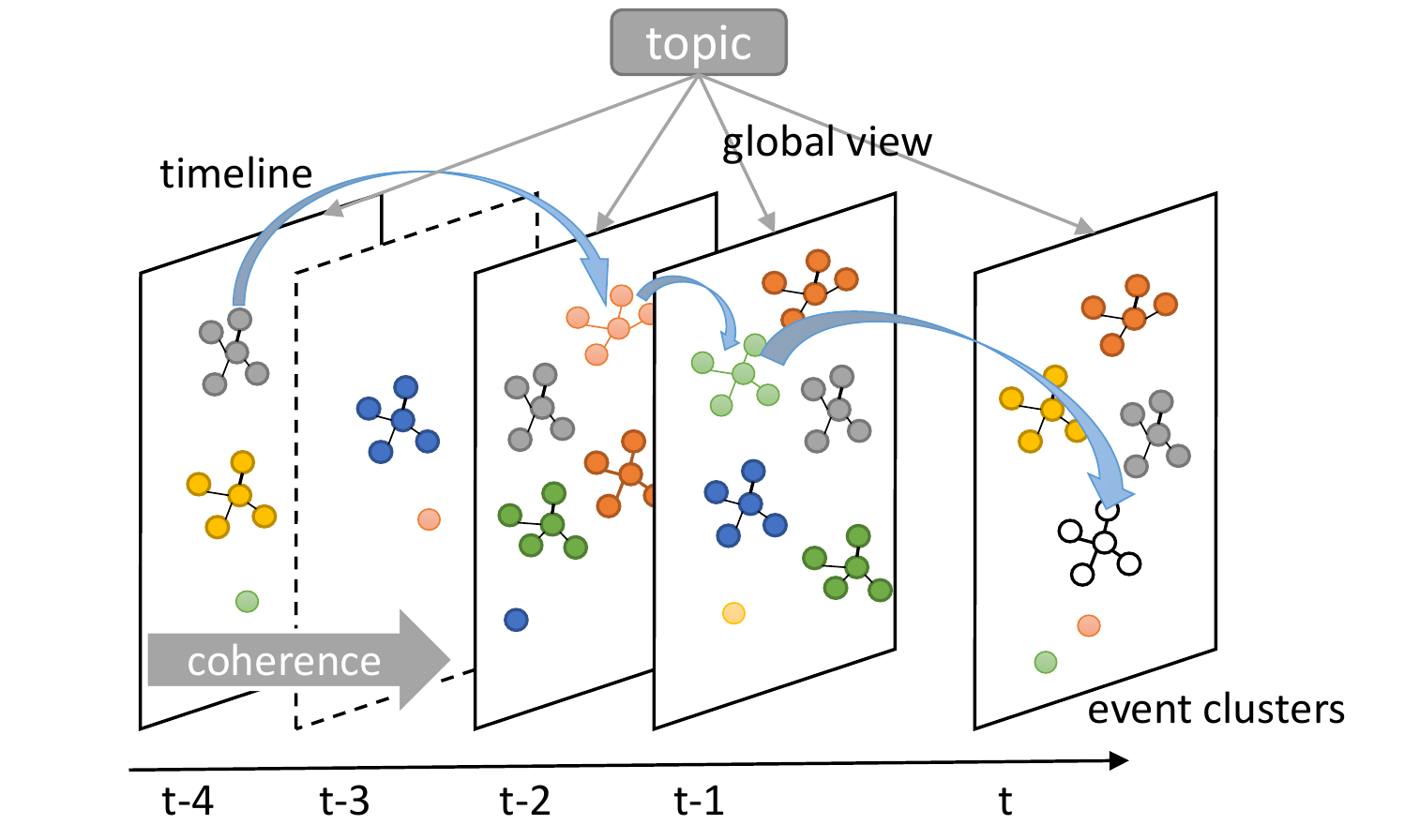}
\caption{Dynamic Evolutionary timeline sequence of event clusters guided by coherence and global view.}
\label{fig:timelineView}
\end{figure}

In reality, there are many repeated reports about the same event in redundant web text.
Hence, taking the paragraphs from $ A $ as a whole, we separate the candidates $ A_i^{j} $  into disjoint clusters in embedding space by affinity propagation algorithm, obtain the number $K$ of clusters $ C = \{C_1, \ldots, C_K \} $, and then divide the clustering result into day by clipping the majority vote of the date. 
The event clusters of $i$-th day and $j$-th class is denoted as:
\begin{equation}
event_{i,j} = \bigcup^{j \in C_j}A_i^{j},\quad 1\leq i \leq T\ and\ 1 \leq j \leq K
\end{equation}
The timeline is the most likely sequence of event clusters, called the Viterbi timeline, shown in Figure ~\ref{fig:timelineView}. We need to find out the dynamic path linking the event clusters as sequence $ TS_{event} $:
\begin{equation}
TS_{event} = \bigcup_{(i,j) \in X} event_{i,j}
\end{equation}
\begin{displaymath}
\displaystyle X = [(t_1, n_1), (t_2, n_2), \ldots \\, (t_{|S|}, n_{|S|})] 
\end{displaymath}
where $ n_i $ is the specific class of event clusters in $ t_i $ day, and $ TS_q $ will be further extracted from $ TS_{event} $.
each embedding of event cluster is defined as the mean value of the candidates $A_i^j $ in the same cluster at vector dimension:
\begin{equation}
v(event_{i,j}) =
\begin{cases}
  \overrightarrow{0}, & \text{if}\ \bigcup^{j \in C_j}A_i^{j} = \emptyset \\
  {\rm AvgPooling} \bigcup^{j \in C_j} v(A_i^{j}), & \text{otherwise}
\end{cases}
\end{equation}

\subsection{Viterbi Timeline}

The Viterbi timeline derives from a origin path $ \displaystyle X' = [(1, n_1), \ldots, \\ (T, n_{T})] $, which is a sequence of clusters $ (i, n_i) $  that generates the event timeline, where $ n_i \in \{C_1, \ldots, C_K\} $.
Two 2-dimensional tables of size 
$ \displaystyle T\times K $ ($W_1$ and $ W_2$) are constructed as follows.

Each element $ \displaystyle W_{1}[i,j] $  of $ \displaystyle W_{1} $  stores the weight of the most likely path so far
$ [n_1, \ldots, n_i] $ with $ n_i=C_j $ that generates event clusters path.
Each element $ \displaystyle W_{2}[i,j] $ of $ \displaystyle W_{2} $ stores $ i-1 $ of the most likely path so far $[(1, n_1), \ldots, (i, n_i=Cj)] $.
The table entries $ \displaystyle W_{1}[i,j]$ and $W_{2}[i,j] $  are filled by increasing order of $ \displaystyle K\cdot i+j $ .
\begin{equation}
W_{1}[i,j]=\max _{k}{(W_{1}[i-1,k]\cdot Q_{kj}^i\cdot R_{ij})}
\end{equation}
\begin{equation}
W_{2}[i,j]=\operatorname{argmax} _{k}{(W_{1}[i-1,k]\cdot Q_{kj}^i\cdot R_{ij})}
\end{equation}
with the dynamic changes of $ k $, $ R_{ij} $ plays the role of global view, and $ Q_{kj}^i $ leverages the adjacent relevance to preserve the coherence, such as defined below. 

\textbf{Coherence}. The size of Transition matrix $ Q $ is $ T\times K\times K $, such that $ Q_{ij}^t $ stores the transition weight of transiting from cluster $ C_i $  to cluster $ C_j $ at the time $t$:
\begin{equation}
Q_{ij}^t = {\rm Cosine}(v(event_{t-1,i}), v(event_{t, j}))
\end{equation}
transition weight is used to measure the relevance of the event clusters between previous and present time status in the consecutive timeline. Timeline is optimally solved by breaking it into sub-relevance and then recursively finding the optimal coherence to the global relevance, which meets the assumption of coherence.

\textbf{Global View}.
The size of Emission matrix $ R $ is $ T\times K $, such that $ R_{ij} $ stores the weight of topic $q$ expression from cluster $C_{j}$ at time $i$:
\begin{equation}
R_{ij} = {\rm Cosine}(v(q), v(event_{i, j}))
\end{equation}
Given a underlying topic, emission weight represents how likely each possible event cluster is along with timeline, which impacts the expression of events in global view and generates the thread running through many of these events.

Temporal association of timeline are the temporal constraining local correlation, and the timeline takes a global view of the related events.
Local correlation $ Q $ and global view $ R $ are interrelated with each other. 
The entire procedure is summarized in Algorithm \ref{alg:generator}.
\begin{algorithm}[htb]
\caption{Capturing the origin path $ X' $}
\label{alg:generator}
\SetKwProg{generate}{Function \emph{generate}}{}{end}
\generate{Viterbi timeline}{
    \ForEach{{\rm cluster} $j \in \{1,\ldots , K \}$}{
        $ W_1[1,j] \leftarrow {\rm Cosine}(v(q), v(event_{1,j}) ) $\;
        $ W_2[1,j] \leftarrow 0 $\;
    }
    \ForEach{ {\rm date} $i \in \{1, \ldots, T  \} $ }{
        $ \alpha = \max_j \max_k Q_{kj}^i \circ R_{kj} $\;
        $ c_v = \frac{\sigma(v(A_i)^{\mathrm{T}
        }v(q))}{\mu(v(A_i)^{\mathrm{T}
        }v(q))}$ \;
        \nlset{(A)}\label{case:A}\If{$\alpha < \beta_{\alpha}$ or $ c_v > \beta_{c_v} $}{
        \ForEach{{\rm cluster} $j \in \{1,\ldots , K \}$}{
            $ W_{1}[i, j] \leftarrow W_{1}[i-1,j] $\;
            $ W_{2}[i,j] \leftarrow -1 $\;
            update $Q_{:,j}^{i+1} $ for $ event_{i,j} = \bigcup^{j \in C_j}A_{i-1}^{j}$\;
        }
        continue\;
        }
        \ForEach{{\rm cluster} $j \in \{1,\ldots , K \}$}{
        $ W_{1}[i,j] \leftarrow \max _{k}{(W_{1}[i-1,k]\cdot Q_{kj}^i\cdot R_{ij})} $\;
        $ W_{2}[i,j] \leftarrow \operatorname{argmax} _{k}{(W_{1}[i-1,k]\cdot Q_{kj}^i\cdot R_{ij})} $\;
        }
    }
    $ z_{T} \leftarrow \operatorname{argmax}_{k}{(W_{1}[T,k])} $\;
    $ n_T \leftarrow C_{z_{T}} $\;
    \For{$ i \leftarrow T,T-1,\ldots, 2 $}{
    $ z_{i-1} \leftarrow W_2[i,z_i] $\;
    $ n_{i-1} \leftarrow C_{z_{i-1}} $\;
    }
    \Return $X'$
}
\end{algorithm}

\subsection{Constraints}
Since the timeline is not necessarily continuous by days, some time-windows may be filled with the event clusters irrelevant to the timeline.
In addition, the event will be reported again by many medias after its first burst, bringing some repeated reports. We propose the operation~\ref{case:A} in Algorithm \ref{alg:generator} to reduce the impact of irrelevance and repeat. 

Relevance $\alpha $ is a measure of continuity at current day, denotes elementwise multiplication of two matrices, and coefficient of variation\footnote{\url{https://en.wikipedia.org/wiki/Coefficient_of_variation}.} $c_v$ is a standardized measure of the burst of the news at that day.
If the adjacent relevance $\alpha $ is too weak, it means that all events at current step can not properly undertake the above information.
With the benchmark of $v(q)$, each news is treated as a sample value at the distribution, and the $c_v$ gives a degree of dispersion about the collected news at that step. 
As showed in operation~\ref{case:A}, we rely on experience to set the hyper-parameters $ \beta_{\alpha}$ and $ \beta_{c_v} $. Once the conditions are met, the previous state of $ W_1 $ and $ W_2 $  will be retained, The $W_2 $ path will be filled with $-1$ to represent jumping over this step. To address the final TS, we filter $ (i, n_i=-1) \in X'$ to get the Viterbi timeline $X$, and extract each $ A_{t_i}^{m_i} $ to generate $TS_q$:
\begin{equation}
m_i = \operatorname{argmax}_{j}{ {\rm Cosine}(v(q), v(A_{t_i}^j))}, \quad A_{t_i}^j \in event_{t_i, n_i}
\end{equation}

\section{Experiments}
We experiment with two public datasets that have been proposed to investigate the timeline. 

\textbf{17 Timelines}~\cite{tran2013leveraging}. The dataset includes 17 timelines published by the major news agencies, such as CNN, BBC, and NBC News.
They developed from 9 different topics, including BP Oil, Michael Jackson Death, H1N1, Haiti Earthquake, Financial Crisis, Libyan War, Iraq War and Egyptian Protest.
Each timeline has its own independent documents set related to the corresponding topic, 
and It overall contains 4,650 news documents of which the timestamps are explicit dates, such as 07 July 2011.

\textbf{Crisis data}~\cite{tran2015timeline}. It includes four crisis topics (wrt. Egypt, Libya, Yemen, Syria), and each topic has around 4,000+ documents with date timestamps. 
The timeline under the same topic has the same document set, which consist of the content and headline of the news articles.
There are totally 25 manually created timelines for these topics.
The headlines are the best summarization for the news, so we focus on headlines timeline on this dataset, rather than extract
sentences from the news documents as candidates.

We compare our proposed framework with these baselines:
\begin{itemize}[wide]
  \item \textbf{Random}: sentences are randomly selected as TS.
  \item \textbf{Chieu et al.}~\cite{chieu2004query}:a multi-document summarizer which utilizes the popularity of a sentence as TF-IDF similarity with other sentences to estimate its importance.
  \item \textbf{ETS}~\cite{yan2011timeline}: a unsupervised TS system in news domain.
  \item \textbf{Tran et al.}~\cite{tran2013leveraging}: a system based on learning to rank techniques, the earliest baseline reported on the 17 Timelines dataset.
  \item \textbf{Regression}~\cite{wang2016socially}: a supervised regression model to extract sentence as summarization.
  \item \textbf{Wang et al.}~\cite{wang2016low}: a low-rank approximation based approach that leverage the matrix factorization techniques and treat the multi-document extractive summarization task as a sentence recommendation problem.
\end{itemize}

We use common summarization metrics (F-measure of ROUGE)~\cite{yan2011evolutionary} to evaluate the quality of the TS generated by models.
The system summarie would be individually evaluated against all reference summaries for the same topic on Crisis data, but against them for the same topic and news agencies on 17 Timelines.
As the standard evalution of the datasets instructs, we treat each sentence as a candidate on 17 Timelines, and adopt each headline of the news on Crisis data. 
To learn the distributed representations, we use pre-trained word vectors\footnote{\url{https://github.com/facebookresearch/fastText}}, trained on Common Crawl and Wikipedia by fastText tookit~\cite{bojanowski2017enriching}.
Furthermore, each $v(q)$ is only embedded by the name of the topic $q$ as a experimental control.

\begin{table}
  \caption{Performance of models on 17 Timelines}
  \label{tab:17 timelines}
  \begin{tabular}{lccc}
    \toprule
    Methods&ROUGE-1&ROUGE-2&ROUGE-S\\
    \midrule
    Random & 0.128 & 0.021 & 0.026\\
    Chieu et al. & 0.202 & 0.037 & 0.041\\
    ETS & 0.207 & 0.047 & 0.042 \\
    Tran et al. & 0.230 & 0.053 & 0.050\\
    Regression & 0.303 & 0.078 & 0.081\\
    Wang et al. & 0.312 & 0.089 & \textbf{0.112}\\
    \hline
    Ours & \textbf{0.334} & \textbf{0.105} & 0.103 \\
  \bottomrule
\end{tabular}
\end{table}

\begin{table}
  \caption{Performance of models on Crisis data}
  \label{tab:Crisis data}
  \begin{tabular}{lccc}
    \toprule
    Methods&ROUGE-1&ROUGE-2&ROUGE-S\\
    \midrule
    Regression & 0.207 & 0.045& 0.039\\
    Wang et al. & 0.211 & 0.046& 0.040\\
    \hline
    Ours & \textbf{0.268} & \textbf{0.057}& \textbf{0.054}\\
  \bottomrule
\end{tabular}
\end{table}

Table~\ref{tab:17 timelines} show the performance of all models on the 17 Timelines dataset.
We can see that our proposed approach is quite comparable to other state-of-the-art models.
It beats others by a large margin by ROUGE-1 and ROUGE-2.
Chieuet al and ETS gives the lower F score, indicating too much handcraft features to constrain TS is restrictive.
Tran et al and Regression are a typical idea of learning to rank, and Wang cast it as a sentence recommendation problem by matrix factorization.
Thought they utilize the supervised information to help the model learning timeline rules, however, supervised model for timeline generation has its insufficiency that the development of timeline about different topic are many and varied.
Our assumption about timeline is well reflected to achieve an ingenious combination of dynamic evolution and distributed representations in the entire framework.
The best result we obtain demonstrates that 
the novel union of distributed representations can benefit the TS task.

We report the results of our framework and the baselines on Crisis data in Table~\ref{tab:Crisis data}.
Regression method is shown as a strong supervised baseline, and Wang's method is the past state-of-the-art method on this dataset.  
They don't explicitly consider much natural feature of the timeline, such as coherence and overall, thus fails to capture the semantic relations between events in short titles.
We lead throughout the timeline by the coherence of the events and a global view of the topic. 
The same hyper-parameters as 17 Timelines are set, 
and it shows that Viterbi timeline can improve the performance significantly in general.

\section{Conclusions}
In this work, we propose a dynamic evolutionary framework for timeline generation, which addresses concerns over events on both coherence and overall. 
At its heart, we propose the Viterbi timeline, and it actually generate the natural association of TS with distributed representations.
Experiments on 17 Timelines and Crisis data demonstrate the effectiveness of the TS framework.
In the future, we would like to base the past and future contexts to generate a finite event sequence as timeline.

%
\bibliographystyle{ACM-Reference-Format}
\bibliography{sample-base}


\begin{thebibliography}{19}


\ifx \showCODEN    \undefined \def \showCODEN     #1{\unskip}     \fi
\ifx \showDOI      \undefined \def \showDOI       #1{#1}\fi
\ifx \showISBNx    \undefined \def \showISBNx     #1{\unskip}     \fi
\ifx \showISBNxiii \undefined \def \showISBNxiii  #1{\unskip}     \fi
\ifx \showISSN     \undefined \def \showISSN      #1{\unskip}     \fi
\ifx \showLCCN     \undefined \def \showLCCN      #1{\unskip}     \fi
\ifx \shownote     \undefined \def \shownote      #1{#1}          \fi
\ifx \showarticletitle \undefined \def \showarticletitle #1{#1}   \fi
\ifx \showURL      \undefined \def \showURL       {\relax}        \fi
\providecommand\bibfield[2]{#2}
\providecommand\bibinfo[2]{#2}
\providecommand\natexlab[1]{#1}
\providecommand\showeprint[2][]{arXiv:#2}

\bibitem[\protect\citeauthoryear{Althoff, Dong, Murphy, Alai, Dang, and
  Zhang}{Althoff et~al\mbox{.}}{2015}]%
        {althoff2015timemachine}
\bibfield{author}{\bibinfo{person}{Tim Althoff}, \bibinfo{person}{Xin~Luna
  Dong}, \bibinfo{person}{Kevin Murphy}, \bibinfo{person}{Safa Alai},
  \bibinfo{person}{Van Dang}, {and} \bibinfo{person}{Wei Zhang}.}
  \bibinfo{year}{2015}\natexlab{}.
\newblock \showarticletitle{TimeMachine: Timeline generation for knowledge-base
  entities}. In \bibinfo{booktitle}{\emph{In SIGKDD}}. ACM,
  \bibinfo{pages}{19--28}.
\newblock


\bibitem[\protect\citeauthoryear{Bojanowski, Grave, Joulin, and
  Mikolov}{Bojanowski et~al\mbox{.}}{2017}]%
        {bojanowski2017enriching}
\bibfield{author}{\bibinfo{person}{Piotr Bojanowski}, \bibinfo{person}{Edouard
  Grave}, \bibinfo{person}{Armand Joulin}, {and} \bibinfo{person}{Tomas
  Mikolov}.} \bibinfo{year}{2017}\natexlab{}.
\newblock \showarticletitle{Enriching Word Vectors with Subword Information}.
\newblock \bibinfo{journal}{\emph{In TACL}} (\bibinfo{year}{2017}),
  \bibinfo{pages}{135--146}.
\newblock
\showISSN{2307-387X}


\bibitem[\protect\citeauthoryear{Chieu and Lee}{Chieu and Lee}{2004}]%
        {chieu2004query}
\bibfield{author}{\bibinfo{person}{Hai~Leong Chieu} {and}
  \bibinfo{person}{Yoong~Keok Lee}.} \bibinfo{year}{2004}\natexlab{}.
\newblock \showarticletitle{Query based event extraction along a timeline}. In
  \bibinfo{booktitle}{\emph{In SIGIR}}. ACM, \bibinfo{pages}{425--432}.
\newblock


\bibitem[\protect\citeauthoryear{Kenter, Borisov, and de~Rijke}{Kenter
  et~al\mbox{.}}{2016}]%
        {kenter2016siamese}
\bibfield{author}{\bibinfo{person}{Tom Kenter}, \bibinfo{person}{Alexey
  Borisov}, {and} \bibinfo{person}{Maarten de Rijke}.}
  \bibinfo{year}{2016}\natexlab{}.
\newblock \showarticletitle{Siamese cbow: Optimizing word embeddings for
  sentence representations}.
\newblock \bibinfo{journal}{\emph{In ACL}} (\bibinfo{year}{2016}).
\newblock


\bibitem[\protect\citeauthoryear{Kessler, Tannier, Hagege, Moriceau, and
  Bittar}{Kessler et~al\mbox{.}}{2012}]%
        {kessler2012finding}
\bibfield{author}{\bibinfo{person}{Remy Kessler}, \bibinfo{person}{Xavier
  Tannier}, \bibinfo{person}{Caroline Hagege}, \bibinfo{person}{V{\'e}ronique
  Moriceau}, {and} \bibinfo{person}{Andr{\'e} Bittar}.}
  \bibinfo{year}{2012}\natexlab{}.
\newblock \showarticletitle{Finding salient dates for building thematic
  timelines}. In \bibinfo{booktitle}{\emph{ACL}}.
\newblock


\bibitem[\protect\citeauthoryear{Levy and Goldberg}{Levy and Goldberg}{2014}]%
        {levy2014neural}
\bibfield{author}{\bibinfo{person}{Omer Levy} {and} \bibinfo{person}{Yoav
  Goldberg}.} \bibinfo{year}{2014}\natexlab{}.
\newblock \showarticletitle{Neural word embedding as implicit matrix
  factorization}. In \bibinfo{booktitle}{\emph{Advances in neural information
  processing systems}}. \bibinfo{pages}{2177--2185}.
\newblock


\bibitem[\protect\citeauthoryear{Li and Cardie}{Li and Cardie}{2014}]%
        {li2014timeline}
\bibfield{author}{\bibinfo{person}{Jiwei Li} {and} \bibinfo{person}{Claire
  Cardie}.} \bibinfo{year}{2014}\natexlab{}.
\newblock \showarticletitle{Timeline generation: Tracking individuals on
  twitter}. In \bibinfo{booktitle}{\emph{In WWW}}. ACM,
  \bibinfo{pages}{643--652}.
\newblock


\bibitem[\protect\citeauthoryear{Li and Li}{Li and Li}{2013}]%
        {li2013evolutionary}
\bibfield{author}{\bibinfo{person}{Jiwei Li} {and} \bibinfo{person}{Sujian
  Li}.} \bibinfo{year}{2013}\natexlab{}.
\newblock \showarticletitle{Evolutionary hierarchical dirichlet process for
  timeline summarization}. In \bibinfo{booktitle}{\emph{In ACL}}.
  \bibinfo{pages}{556--560}.
\newblock


\bibitem[\protect\citeauthoryear{Liang, Yilmaz, and Kanoulas}{Liang
  et~al\mbox{.}}{2016}]%
        {liang2016dynamic}
\bibfield{author}{\bibinfo{person}{Shangsong Liang}, \bibinfo{person}{Emine
  Yilmaz}, {and} \bibinfo{person}{Evangelos Kanoulas}.}
  \bibinfo{year}{2016}\natexlab{}.
\newblock \showarticletitle{Dynamic clustering of streaming short documents}.
  In \bibinfo{booktitle}{\emph{In SIGKDD}}. \bibinfo{pages}{995--1004}.
\newblock


\bibitem[\protect\citeauthoryear{Liu, Niu, Lai, Kong, and Xu}{Liu
  et~al\mbox{.}}{2017}]%
        {liu2017growing}
\bibfield{author}{\bibinfo{person}{Bang Liu}, \bibinfo{person}{Di Niu},
  \bibinfo{person}{Kunfeng Lai}, \bibinfo{person}{Linglong Kong}, {and}
  \bibinfo{person}{Yu Xu}.} \bibinfo{year}{2017}\natexlab{}.
\newblock \showarticletitle{Growing Story Forest Online from Massive Breaking
  News}. In \bibinfo{booktitle}{\emph{In CIKM}}. ACM,
  \bibinfo{pages}{777--785}.
\newblock


\bibitem[\protect\citeauthoryear{Mikolov, Chen, Corrado, and Dean}{Mikolov
  et~al\mbox{.}}{2013}]%
        {mikolov2013efficient}
\bibfield{author}{\bibinfo{person}{Tomas Mikolov}, \bibinfo{person}{Kai Chen},
  \bibinfo{person}{Greg Corrado}, {and} \bibinfo{person}{Jeffrey Dean}.}
  \bibinfo{year}{2013}\natexlab{}.
\newblock \showarticletitle{Efficient estimation of word representations in
  vector space}.
\newblock \bibinfo{journal}{\emph{arXiv:1301.3781}} (\bibinfo{year}{2013}).
\newblock


\bibitem[\protect\citeauthoryear{Mishra and Berberich}{Mishra and
  Berberich}{2016}]%
        {mishra2016event}
\bibfield{author}{\bibinfo{person}{Arunav Mishra} {and} \bibinfo{person}{Klaus
  Berberich}.} \bibinfo{year}{2016}\natexlab{}.
\newblock \showarticletitle{Event digest: A holistic view on past events}. In
  \bibinfo{booktitle}{\emph{In SIGIR}}. ACM, \bibinfo{pages}{493--502}.
\newblock


\bibitem[\protect\citeauthoryear{Tran, Alrifai, and Herder}{Tran
  et~al\mbox{.}}{2015}]%
        {tran2015timeline}
\bibfield{author}{\bibinfo{person}{Giang Tran}, \bibinfo{person}{Mohammad
  Alrifai}, {and} \bibinfo{person}{Eelco Herder}.}
  \bibinfo{year}{2015}\natexlab{}.
\newblock \showarticletitle{Timeline summarization from relevant headlines}. In
  \bibinfo{booktitle}{\emph{In ECIR}}. Springer, \bibinfo{pages}{245--256}.
\newblock


\bibitem[\protect\citeauthoryear{Tran, Tran, Tran, Alrifai, and Kanhabua}{Tran
  et~al\mbox{.}}{2013}]%
        {tran2013leveraging}
\bibfield{author}{\bibinfo{person}{Giang~Binh Tran}, \bibinfo{person}{Tuan~A
  Tran}, \bibinfo{person}{Nam-Khanh Tran}, \bibinfo{person}{Mohammad Alrifai},
  {and} \bibinfo{person}{Nattiya Kanhabua}.} \bibinfo{year}{2013}\natexlab{}.
\newblock \showarticletitle{Leveraging learning to rank in an optimization
  framework for timeline summarization}. In \bibinfo{booktitle}{\emph{SIGIR
  TAIA}}.
\newblock


\bibitem[\protect\citeauthoryear{Wang, Cardie, and Marchetti}{Wang
  et~al\mbox{.}}{2015}]%
        {wang2016socially}
\bibfield{author}{\bibinfo{person}{Lu Wang}, \bibinfo{person}{Claire Cardie},
  {and} \bibinfo{person}{Galen Marchetti}.} \bibinfo{year}{2015}\natexlab{}.
\newblock \showarticletitle{Socially-informed timeline generation for complex
  events}.
\newblock \bibinfo{journal}{\emph{In NAACL}} (\bibinfo{year}{2015}).
\newblock


\bibitem[\protect\citeauthoryear{Wang, Mehdad, Radev, and Stent}{Wang
  et~al\mbox{.}}{2016}]%
        {wang2016low}
\bibfield{author}{\bibinfo{person}{William~Yang Wang}, \bibinfo{person}{Yashar
  Mehdad}, \bibinfo{person}{Dragomir~R Radev}, {and} \bibinfo{person}{Amanda
  Stent}.} \bibinfo{year}{2016}\natexlab{}.
\newblock \showarticletitle{A low-rank approximation approach to learning joint
  embeddings of news stories and images for timeline summarization}. In
  \bibinfo{booktitle}{\emph{In NAACL}}. \bibinfo{pages}{58--68}.
\newblock


\bibitem[\protect\citeauthoryear{Yan, Kong, Huang, Wan, Li, and Zhang}{Yan
  et~al\mbox{.}}{2011a}]%
        {yan2011timeline}
\bibfield{author}{\bibinfo{person}{Rui Yan}, \bibinfo{person}{Liang Kong},
  \bibinfo{person}{Congrui Huang}, \bibinfo{person}{Xiaojun Wan},
  \bibinfo{person}{Xiaoming Li}, {and} \bibinfo{person}{Yan Zhang}.}
  \bibinfo{year}{2011}\natexlab{a}.
\newblock \showarticletitle{Timeline generation through evolutionary
  trans-temporal summarization}. In \bibinfo{booktitle}{\emph{In EMNLP}}.
  \bibinfo{pages}{433--443}.
\newblock


\bibitem[\protect\citeauthoryear{Yan, Wan, Otterbacher, Kong, Li, and
  Zhang}{Yan et~al\mbox{.}}{2011b}]%
        {yan2011evolutionary}
\bibfield{author}{\bibinfo{person}{Rui Yan}, \bibinfo{person}{Xiaojun Wan},
  \bibinfo{person}{Jahna Otterbacher}, \bibinfo{person}{Liang Kong},
  \bibinfo{person}{Xiaoming Li}, {and} \bibinfo{person}{Yan Zhang}.}
  \bibinfo{year}{2011}\natexlab{b}.
\newblock \showarticletitle{Evolutionary timeline summarization: a balanced
  optimization framework via iterative substitution}. In
  \bibinfo{booktitle}{\emph{In SIGIR}}. ACM, \bibinfo{pages}{745--754}.
\newblock


\bibitem[\protect\citeauthoryear{Zhou, Guo, and He}{Zhou et~al\mbox{.}}{2018}]%
        {zhou2018neural}
\bibfield{author}{\bibinfo{person}{Deyu Zhou}, \bibinfo{person}{Linsen Guo},
  {and} \bibinfo{person}{Yulan He}.} \bibinfo{year}{2018}\natexlab{}.
\newblock \showarticletitle{Neural Storyline Extraction Model for Storyline
  Generation from News Articles}. In \bibinfo{booktitle}{\emph{In NAACL}}.
  \bibinfo{pages}{1727--1736}.
\newblock


\end{thebibliography}

%

\end{document}